\documentclass[10pt,twocolumn,letterpaper]{article}

\usepackage{iccv}
\usepackage{times}
\usepackage{epsfig}
\usepackage{graphicx}
\usepackage{amsmath}
\usepackage{amssymb}
\usepackage{hyphenat}
\usepackage[table]{xcolor}
\usepackage[font=small,belowskip=0pt, aboveskip=6pt]{caption}
\usepackage{multirow}
\usepackage{booktabs,amsfonts,dcolumn}
\usepackage{tabularx}
\usepackage{makecell}
\usepackage{hyperref}
\usepackage{comment}
\usepackage{diagbox}
\usepackage[style=ieee,mincitenames=1,maxcitenames=2,maxbibnames=10,doi=false,isbn=false,url=false,eprint=false]{biblatex}
\newcommand\fig[1]{Fig.~\ref{#1}}

\makeatletter
\DeclareRobustCommand\onedot{\futurelet\@let@token\@onedot}
\def\@onedot{\ifx\@let@token.\else.\null\fi\xspace}

\def\eg{\emph{e.g}\onedot}

\def\etal{\emph{et al}\onedot}
\makeatother

\iccvfinalcopy 

\def\iccvPaperID{****} 

\begin{document}

\title{\LARGE \bf
CaveSeg: Deep Semantic Segmentation and Scene Parsing for Autonomous Underwater Cave Exploration}

\author{Adnan Abdullah$^{1}$, Titon Barua$^{2}$, Reagan Tibbetts$^{2}$, Zijie Chen$^{1}$, Md Jahidul Islam$^{1}$, and Ioannis Rekleitis$^{2}$ \\
{\tt\small adnanabdullah@ufl.edu, baruat@email.sc.edu, rbt@email.sc.edu,}  \\ 
{\tt\small chen.zijie@ufl.edu, jahid@ece.ufl.edu, yiannisr@cse.sc.edu} \\
{
\small $^{1}$RoboPI Laboratory, Department of ECE, University of Florida, FL 32611, USA.} \\ 
{
\small $^{2}$AFRL Laboratory, Department of CSE, University of South Carolina, SC 29208, USA. }
{
\thanks{This pre-print is accepted for publication at ICRA 2024.} \thanks{Project: \url{https://robopi.ece.ufl.edu/caveseg.html}}
}
}


\maketitle

\begin{abstract}
In this paper, we present CaveSeg - the first visual learning pipeline for semantic segmentation and scene parsing for AUV navigation inside underwater caves. We address the problem of scarce annotated training data by preparing a comprehensive dataset for semantic segmentation of underwater cave scenes. It contains pixel annotations for important navigation markers (\eg caveline, arrows), obstacles (\eg ground plane and overhead layers), scuba divers, and open areas for servoing. Through comprehensive benchmark analyses on cave systems in USA, Mexico, and Spain locations, we demonstrate that robust deep visual models can be developed based on CaveSeg for fast semantic scene parsing of underwater cave environments. In particular, we formulate a novel transformer-based model that is computationally light and offers near real-time execution in addition to achieving state-of-the-art performance. Finally, we explore the design choices and implications of semantic segmentation for visual servoing by AUVs inside underwater caves. The proposed model and benchmark dataset open up promising opportunities for future research in autonomous underwater cave exploration and mapping.
\end{abstract}

\vspace{-3mm}
\section{Introduction and Background}

\begin{figure}[t]
     \centering
     {\includegraphics[width=\columnwidth]{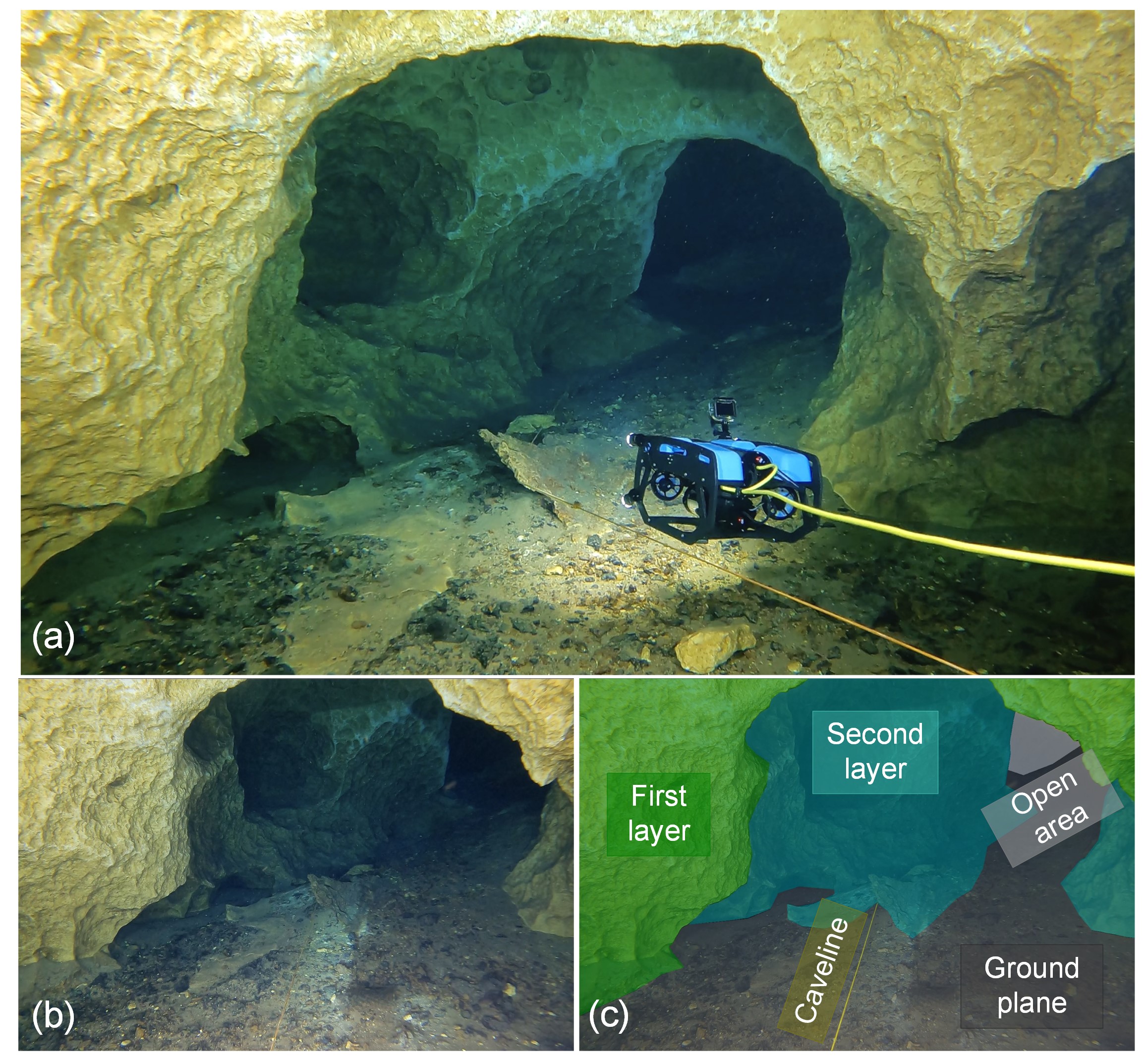}}%
     \vspace{-2mm}
     \caption{(a) A tethered BlueROV2 is operating inside Orange Grove underwater cave system in FL, USA; it is teleoperated by a surface operator  following the \emph{caveline} as a navigation guide; (b) the corresponding POV from the robot's camera; (c) the proposed semantic parsing concept is shown; the envisioned capabilities are: first-layer \& second-layer obstacle avoidance, ground plane estimation, and caveline detection, following, and 3D estimation -- to enable autonomous robot navigation inside underwater caves.}%
     \vspace{-6mm}
     \label{fig:beautyshot}
 \end{figure}
 
Underwater cave formations, sediment properties, and water chemistry provide insights into the world's past climate conditions and geological processes~\cite{gonzalez2008arrival,richmond2020autonomous}. Underwater caves also play a crucial role in monitoring and tracking groundwater flows in Karst topographies, while almost $25\%$ of the world's population relies on Karst freshwater resources~\cite{karstbook}. Despite the importance, underwater cave exploration and mapping by humans is a tedious, labor-intensive, and extremely dangerous operation, even for highly skilled divers~\cite{buzzacott2009american}. When a new section of a cave is discovered, a single and continuous line termed \emph{caveline}~\cite{exley1986basic} is laid out identifying the skeleton of the main passages. The caveline is attached to other navigation guides such as \emph{arrows} and \emph{cookies}~\cite{yu2023weakly}, marking the orientation of the cave, distance to the entrance and presence of other divers. Such surveys by the explorers produce a one-dimensional retraction of the 3D environment. Recording all this information together with additional observations~\cite{Burge1988Survey} is a challenging, time-consuming, and error-prone process.

Enabling Autonomous Underwater Vehicles (AUVs) and Remotely Operated Vehicles (ROVs) to navigate, explore, and map underwater caves safely and effectively is of significant importance~\cite{joshi2022underwater,richmond2020autonomous}; \fig{fig:beautyshot}(a) shows an ROV deployment scenario inside the Orange Grove Cave System in Florida. Our earlier work~\cite{JoshiICRA2022} developed a high-precision camera trajectory estimation method by a Visual-Inertial Odometry (VIO) algorithm~\cite{RahmanIJRR2022}, which can generate 3D caveline trajectory estimates that are comparable to the manually surveyed measurements. More recently in~\cite{yu2023weakly}, we addressed the lack of annotated samples in visual learning for caveline detection and tracking across different scenes for AUV navigation.

In this work, we focus on developing a deep visual learning pipeline to extract dense semantic information for autonomous underwater cave exploration by mobile robots. Considering the limited onboard resources available on embedded platforms, our objective is to design a computationally light model that can (learn to) identify the navigation markers of underwater caves (\eg caveline, arrows), obstacles to avoid (\eg ground plane and overhead layers), scuba divers (for cooperative missions), and safe open areas for visual servoing in real-time. We identify two major difficulties to achieve these: (\textbf{i}) no large-scale datasets are available for underwater cave environments; (\textbf{ii}) the state-of-the-art (SOTA) models for semantic scene parsing are computationally too demanding for robotic platforms.

\begin{figure*}[t]
     \centering
     {\includegraphics[width=\linewidth]{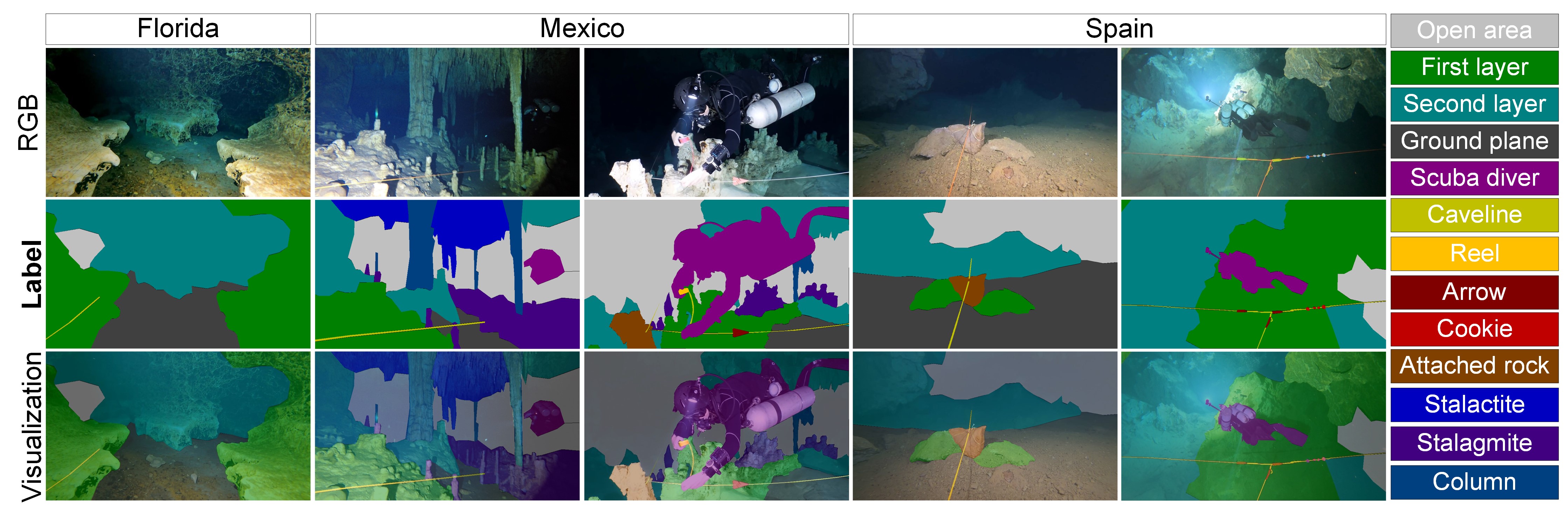}}%
     \vspace{-3mm}
     \caption{A few sample images from the proposed \textbf{CaveSeg dataset}, corresponding ground truth labels, and their overlayed visualizations are shown; color codes for each object category are listed on the right.}%
     \label{fig:dataset}
 \end{figure*}

We address these challenges by proposing \textbf{CaveSeg}, the first large-scale semantic segmentation dataset and learning pipeline for underwater cave exploration. We collect comprehensive training data by ROVs and scuba divers through robotics trials in three major locations~\cite{yu2023weakly}: the Devil's system in Florida, USA; Dos Ojos Cenote in QR, Mexico; and Cueva del Agua in Murcia, Spain. Our processed data contain $3350$ pixel-annotated samples with $13$ object categories that include first and second layer obstacles, human scuba divers, and navigation aids (see Sec~\ref{data_sec}; Fig.~\ref{fig:dataset}). We also compile a \textbf{CaveSeg-Challenge} test set that contains $350$ samples from unseen waterbody and cave systems such as the Blue Grotto and Orange Grove cave systems in FL, USA. We conduct extensive benchmark evaluation of SOTA models across the Convolutional Neural Network (CNN)~\cite{girshick2014rich,zhang2018context}, Conditional Random Fields (CRF)~\cite{chen2014semantic,chen2017deeplab}, and Vision Transformer (ViT)~\cite{xie2021segformer,strudel2021segmenter,liu2021swin} literature, which validate that robust semantic learning is feasible on CaveSeg.      

Moreover, we develop a novel \textbf{CaveSeg model} by rigorous design choices to balance the robustness-efficiency trade-off. The proposed model consists of a transformer-based backbone, a multi\hyp scale pyramid pooling head, and a hierarchical feature aggregation module for robust semantic learning (see Sec.\ref{model}). Experimental evaluation and comparison with SOTA models suggest that the CaveSeg model is over $3\times$ more memory efficiency and offers $1.8\times$ faster inference than SOTA models, while providing comparable benchmark performances. A series of experiments on unseen challenging scenes prove the robustness of our model across different waterbody conditions and optical artifacts; see more details in Sec.\ref{perform_sec}.

Furthermore, we highlight several challenging scenarios and practical use cases of CaveSeg for real-time AUV navigation inside underwater caves. Those scenarios include safe cave exploration by caveline following and obstacle avoidance, planning towards caveline rediscovery, finding safe open passages and exit directions inside caves, giving uninterrupted right-of-way to scuba divers exiting the cave, and 3D semantic mapping and state estimation. We demonstrate that CaveSeg-generated semantic labels can be utilized effectively for vision-based cave exploration and semantic mapping by AUVs (see more in Sec.~\ref{use_cases}).


\section{Related Work}

\subsection{Underwater Cave Exploration and Mapping}
Exploration and mapping of underwater caves by human divers traditionally employed  photogrammetry~\cite{fortin2021environmental} methods in order to generate informative photorealistic representation, especially for sites with archaeological interest~\cite{rissolo2015novel,gonzalez2008arrival}. Autonomous underwater cave mapping by AUVs remains an open problem due to many challenges. Mallios \etal~\cite{mallios2016toward} manually moved an AUV collecting acoustic data for offline mapping, and Weidner \etal~\cite{WeidnerICRA2017,WeidnerMSc2017} used images from a stereo camera to create a 3D reconstruction of the cave walls, floor, and ceiling. Major challenges to vision\hyp based  state estimation in an underwater environment include lighting variations, light absorption, and blurriness~\cite{JoshiIROS2019}. Rahman \etal~\cite{RahmanICRA2018,RahmanIROS2019a,RahmanIJRR2022} proposed a framework where visual, acoustic,  inertial, and water depth data are fused together to estimate the trajectory of an AUV or a sensor while in parallel generates a sparse representation of the underwater cave. Denser models of the cave boundaries can be obtained by mapping the moving shadows~\cite{RahmanIROS2019b}, the contours~\cite{massone2020contour}, or dense online stereo reconstruction~\cite{WangICRA2023}. 

More recently, Richmond~\etal~\cite{richmond2020autonomous} describe a man\hyp portable AUV named \emph{Sunfish}, which can work safely inside underwater caves and bring back chemical profiles, detailed imagery, and sonar maps of the cave. Our recent work~\cite{yu2023weakly} develops a ViT-based caveline detection model for autonomous caveline following by underwater robots. However, beyond detecting cavelines, a full-form scene parsing is essential for safe and effective AUV navigation inside underwater caves. 

\subsection{Semantic Segmentation of Underwater Scenes}
Deep visual learning algorithms have revolutionized semantic segmentation and scene parsing benchmarks on standard datasets, which are mostly developed for terrestrial applications. The existing learning pipelines trained on terrestrial imagery are not directly applicable because the object categories and image statistics are entirely different. A unique set of underwater image distortion artifacts and the unavailability of large-scale annotated datasets have influenced a significant lack of research attempts on semantic segmentation of underwater imagery. The contemporary research literature offers application-specific image datasets for coral reef segmentation and coverage estimation~\cite{alonso2019coralseg,ravanbakhsh2015automated}, visual attention modeling~\cite{islam2022svam}, human-robot cooperative missions~\cite{islam2020suim}, image restoration and foreground enhancement~\cite{islam2020sesr}, and fish detection~\cite{alshdaifat2020improved,ulucan2020large}. 

\begin{figure}[h]
\vspace{-1mm}
     \centering     {\includegraphics[width=\columnwidth]{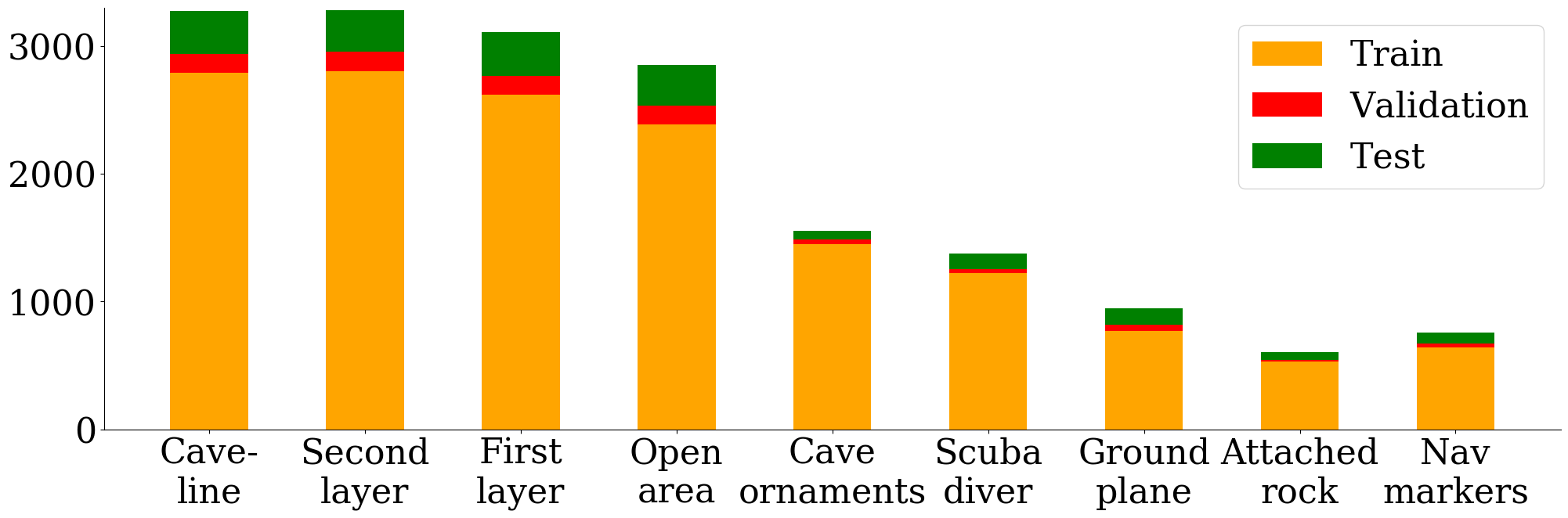}}%
     \vspace{-1mm}
     \caption{Frequencies and distributions of important object categories are shown in the \textit{train}, \textit{validation}, and \textit{test} sets.}%
     \vspace{-1mm}
     \label{fig:data_stat}
 \end{figure}

More recent work by Modasshir~\etal combined a deep learning-based classifier model to identify and track the locations of different types of corals to generate semantic maps~\cite{ModasshirRobio2018} as well as volumetric models~\cite{ModasshirFSR2019}. Islam~\etal formulated the SUIM dataset~\cite{islam2020suim} for semantic segmentation of underwater imagery with eight object categories: fish, coral reefs, aquatic plants, wrecks/ruins, human divers, robots/instruments, and sea-floor. Other datasets consider even fewer object categories such as marine debris or ship hull defects~\cite{waszak2022semantic}. With limited training samples per object category over only a few waterbody types, it is extremely challenging to achieve good generalization performance by SOTA deep learning-based models for image segmentation and scene parsing. More importantly, these object categories are not useful for underwater cave exploration and mapping applications -- which we address in this paper. 

\section{CaveSeg Dataset: Data Preparation and Problem Formulation}\label{data_sec}

\begin{figure*}[t]
     \centering
     {\includegraphics[width=\linewidth]{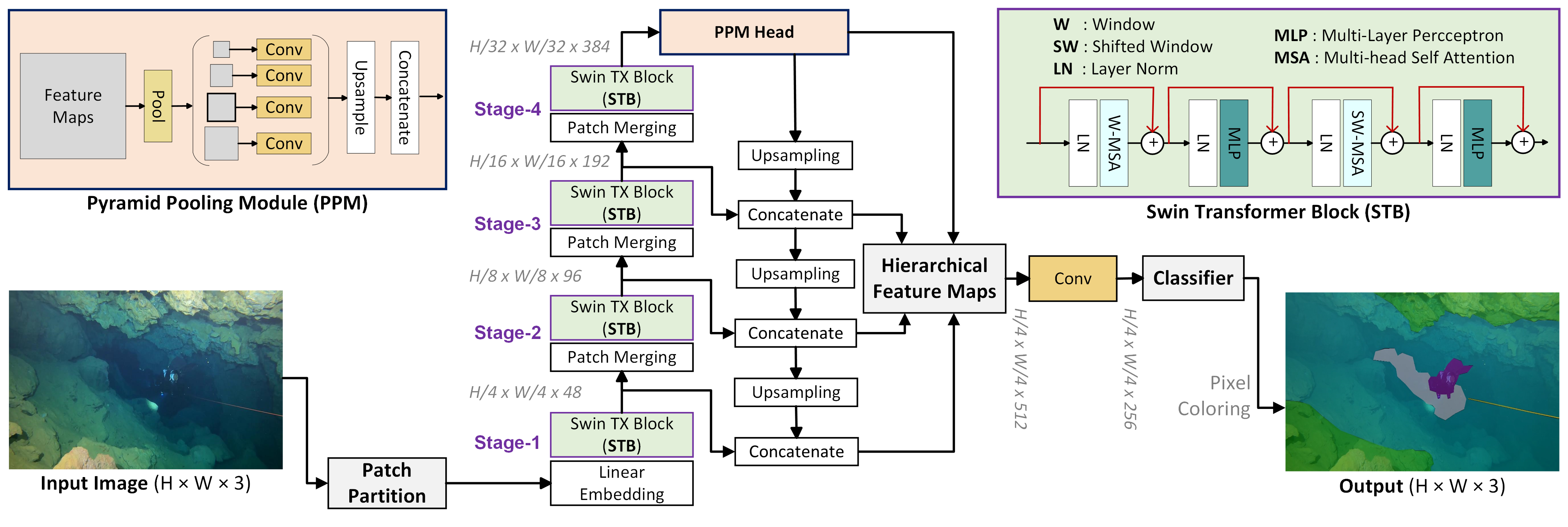}}%
     \vspace{-2mm}
     \caption{The network architecture of our proposed \textbf{CaveSeg model} is shown. Input images are partitioned into $4\times4$ patches and fed into a four-stage transformer backbone for coarse-to-fine feature extraction. The extracted multi-scale features are then pooled and combined by the PPM head for bottom-up and top-down feature aggregation. A hierarchical feature map is then compiled by merging several multi-level feature representations. On this feature space, a classifier performs pixel-level semantic segmentation to generate the final outputs.}%
     \label{fig:model}
     \vspace{-3mm}
 \end{figure*}

We prepared learning pipelines with data collected from \textbf{three cave systems} in different geographical locations~\cite{yu2023weakly}: the Devil's system in Florida, USA; Dos Ojos Cenote in QR, Mexico; and Cueva del Agua in Murcia, Spain. For the semantic labels, we considered the following object categories: caveline, first layer (immediate avoidance areas), second layer (areas to be avoided subsequently), open area (obstacle-free regions), ground plane, scuba divers, navigation aids (\eg, arrows, reels, and cookies), and caveline-attached rocks. Moreover, cave ornaments (\eg, stalactites, stalagmites, and columns) are also considered if they are present in the scene, which is the case for the Mexico caves. With these $13$ object categories in consideration, a total of $3350$ images are labeled in the \textbf{CaveSeg dataset}. A few annotated samples are shown in Fig.~\ref{fig:dataset}; the dataset and relevant information are available online at \url{https://robopi.ece.ufl.edu/caveseg.html}. 

As mentioned earlier, the role of the caveline is crucial for any underwater cave operations. It represents the direction of exploration, the main area where the cave extends, and equally important, the path to safely exit the cave. Caveline pixels are marked as yellow in order to generate maximum contrast. The rock formations where the caveline is attached are often called \emph{placements} or \emph{attachment points}. These rocks most of the time signal a change in the direction of the caveline and they are marked in brown. Navigational markers such as \emph{arrows} (dark red) and \emph{cookies} (red) provide important information about the direction to the nearest cave exit and the presence of other divers in the cave, respectively.

A special category in our semantic mapping scheme is the scuba divers (magenta). An AUV inside the cave should always give the right of way to divers, especially those exiting the cave. In case of an emergency, there should be nothing impeding the divers from reaching the surface, which is a norm practiced by cave divers. As such we have established a human diver category to incorporate emergency responses when a diver is detected, for example, lowering the light intensity, moving away from the main passage, avoiding abrupt motions, etc. 

Fig.~\ref{fig:data_stat} shows the frequency and distribution of various object categories in CaveSeg dataset. Human diver is present
in $40$\% of the images. Over $90\%$ of the samples contain caveline, obstacle-free open areas, as well as the first and second layer obstacles. Three types of cave ornaments are frequently seen in the Mexico caves and jointly they are present in $40\%$ of the data. Navigation markers (\eg, cookies, arrows, reels) typically occupy smaller pixel areas compared to other objects, and they are found in $20\%$ of the data. The training set is carefully compiled to represent adequate samples for each category while ensuring a unique and unseen challenge set for testing. Overall, these $13$ object categories in the scene embed useful information for vision-based planning and navigation for AUVs inside underwater caves. 

\section{CaveSeg Model: Semantic Scene Segmentation Of Underwater Caves}\label{model}
\subsection{Network Design and Learning Pipeline} 
We search for a computationally light architecture that provides real-time underwater cave scene segmentation in addition to achieving state-of-the-art (SOTA) performance.
To this end, we explore both CNN-based and transformer-based backbone architectures~\cite{chen2018encoder,wu2019fastfcn,xie2021segformer}. The windowed multi-head self-attention (W-MSA) module proposed in Swin Transformer~\cite{liu2021swin} is a powerful tool to maintain efficient computation. Additionally, the window shifting technique connects features across spatially overlapped windows. We find this connection to be particularly useful since cave scenes hold some spatial relation among categories. For instance, navigation markers and attachment rocks are almost always found on the caveline, while the second layer obstacles are usually found around the first layer or ground plane. Although computationally heavy, we found such cross-category attention extraction to be effective in cave scene segmentation tasks. Inspired by this, we design a novel architecture that incorporates these capabilities with a light Swin Transformer backbone for feature extraction. With an efficient Pyramid Pooling Module (PPM) and hierarchical feature aggregation, our proposed CaveSeg model is over $3.3\times$ memory efficient and offers $60\%$ faster inference rates than the {Swin Transformer} base model.  

\subsubsection{CaveSeg Model Architecture}
The detailed network architecture is shown in Fig~\ref{fig:model}. First, input RGB images are partitioned into $4\times4$ non-overlapping
patches or {\tt tokens}; a linear embedding layer of dimension $48$ is then applied to each token. These feature tokens are passed through a four-stage backbone, each containing a windowed and shifted windowed module of multi-head self-attention~\cite{vaswani2017attention}. In each stage, patches are merged with $2\times2$ neighboring regions to reduce the number of tokens while at the same time, the linear embedded dimension is doubled. 
Subsequently, bottom-up and top-down feature aggregation~\cite{lin2017feature} is performed in two separate branches. A pyramid pooling module (PPM)~\cite{zhao2017pyramid} is attached to the backbone that further improves global feature extraction at deep layers of the network~\cite{xiao2018unified}. Features from each stage of the backbone as well as from the PPM head are then fused and compiled into a hierarchical feature map. Finally, a fully connected convolution layer performs pixel-wise category estimation on that feature space. The {\tt token size}, {\tt embed dimension}, and {\tt neighbor size} are empirically set to maximize memory efficiency at the lowest possible accuracy cost. Other parameters of shifted-window module and PPM module are directly used from their original architecture.

\subsubsection{Supervised Learning Pipeline}
The end-to-end training is driven by the standard cross-entropy loss~\cite{krizhevsky2012imagenet} that quantifies the dissimilarity in the generated and predicted pixel labels for each category. For multi-class segmentation of an image with $N$ pixels and $M$ classes, it is calculated as
\begin{equation}
{\mathcal L}_{CE} = -\frac{1}{N}\sum_{i=1}^N\sum_{c=1}^My_{i,c}\log(p_{i,c}),
\end{equation}
where $p_{i,c}$ denotes the probability of pixel $i$  belonging to class $c$, and $y$ is $0$ (or $1$) for correct (or incorrect) predictions, respectively. The training is optimized by the Stochastic Gradient Descent (SGD) algorithm~\cite{robbins1951stochastic} with an initial learning rate of $1e^{-4}$ and a momentum of $0.9$. 

\subsection{Training Setups for CaveSeg \& Baseline Models}
We configured a unified training pipeline for CaveSeg and several SOTA benchmark models for training and evaluation. Specifically, we used the MMSegmentation libraries~\cite{mmseg2020} in PyTorch for the large-scale training on CaveSeg dataset with a $85$:$5$:$10$ split ratio for train, validation, and test, respectively. For baseline comparison, we considered the following SOTA models across the CNN, CRF, and ViT literature: FastFCN~\cite{wu2019fastfcn}, DeepLabV3+~\cite{chen2018encoder}, Segmenter~\cite{strudel2021segmenter}, Segformer~\cite{xie2021segformer}, and Swin Transformer~\cite{liu2021swin}. The proposed CaveSeg model is pre-trained on ADE20k~\cite{zhou2017scene} followed by the unified training. The input-output resolution is set to $960\times540$ for all models; other SOTA model-specific parameters are chosen based on their recommended configurations.

\begin{table}[h]
\vspace{-2mm}
        \centering
    \caption{Semantic segmentation performances of all models are compared on $350$ test images from the CaveSeg-Challenge set; here, higher scores ($\uparrow$) are better for all metrics in consideration.
    }
    \vspace{-1mm}
    \begin{tabular}{l|r|r|r}
    \Xhline{2\arrayrulewidth}
         \textbf{Method}  & \textbf{mIoU} $\uparrow$ & \textbf{mAcc} $\uparrow$ & \textbf{aAcc} $\uparrow$ \\
         \Xhline{2\arrayrulewidth}
         FastFCN~\cite{wu2019fastfcn} & $38.86$ & $46.89$ & $72.01$  \\
          DeepLabV3+~\cite{chen2018encoder} & $38.46$ & $49.47$ & $71.64$  \\
         Segmenter~\cite{strudel2021segmenter} & $30.81$ & $39.64$ & $69.76$ \\
         Segformer~\cite{xie2021segformer} & $35.36$ & $44.71$ & $70.19$ \\
         Swin Transformer~\cite{liu2021swin} & $\textbf{48.11}$ & $\textbf{56.69}$ & $\textbf{73.26}$ \\
         \textbf{CaveSeg} (proposed)  & $40.22$ & $45.99$  & $72.91$ \\
         \Xhline{2\arrayrulewidth}
    \end{tabular}
    \label{tab:result}
    \vspace{-2mm}
\end{table}

\section{Performance Analyses of CaveSeg}\label{perform_sec}

\begin{figure*}[t]
     \centering
     {\includegraphics[width=\linewidth]{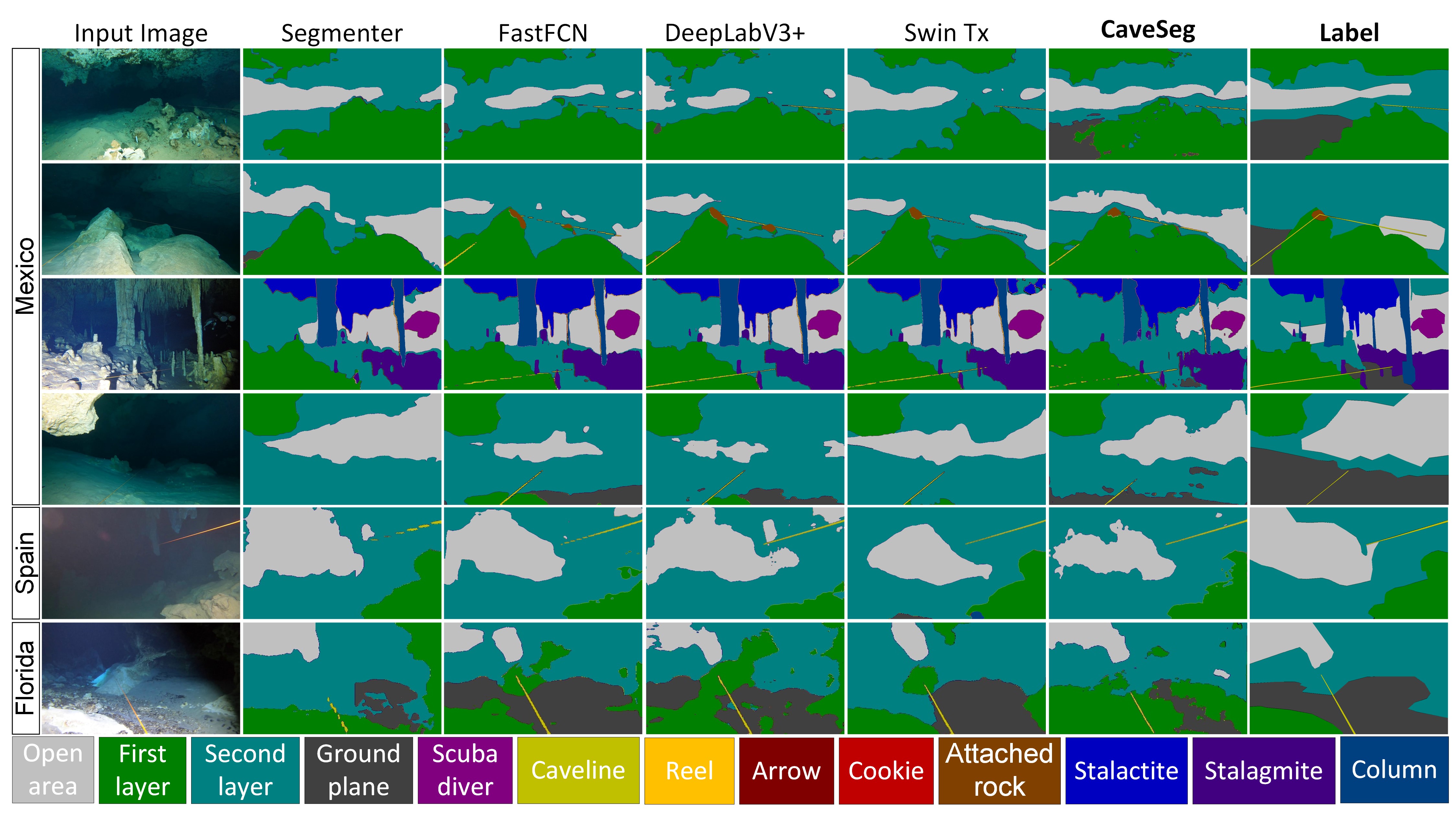}}%
     \vspace{-2mm}
     \caption{A few qualitative performance comparisons of all models on CaveSeg-Challenge test set are shown (results for only five top-performing models are shown for clarity). Note that the object detection and localization accuracy for categories such as caveline, open area, and navigation markers are particularly important for AUV navigation.}%
     \label{fig:comparison}
     \vspace{-2mm}
 \end{figure*}
 
\subsection{Quantitative Evaluation}
We use three standard metrics~\cite{long2015fully,minaee2021image} for quantitative assessments: mean Intersection Over Union \textbf{mIOU}, mean class-wise Accuracy (\textbf{mAcc}), and Average pixel Accuracy (\textbf{aAcc}). The {IoU} ({Intersection Over Union)} measures caveline localization performance using the area of overlapping regions of the predicted and ground truth labels. it is defined as $IoU = \frac{Area\text{ }of\text{ }overlap}{Area\text{ }of\text{ }union}$. On the other hand, mAcc and aAcc represent the mean accuracy of each class category and over all pixels, respectively. The quantitative results are presented in Table~\ref{tab:result}; all evaluations are performed on the \textbf{CaveSeg-Challenge} set, which we curated with $350$ test samples. These samples include low-light noisy scenarios compiled in the CL-Challenge set (released earlier in~\cite{yu2023weakly}), as well as data from cave explorations in other such as the Blue Grotto and Orange Grove cave systems in FL, USA.

As Table~\ref{tab:result} shows, our proposed dataset and learning pipelines can achieve up to $73\%$ accuracy for pixel-wise segmentation. The proposed CaveSeg model offers comparable performance margins despite having a significantly lighter architecture. The comparisons for computational efficiency are provided in Table~\ref{tab:efficiency}. CaveSeg model has less than $50\%$ parameters than other competitive models and offers up to $1.8\times$ faster inference rates. While we analyze the class-wise performance, we find that small and rare object categories such as arrows, cookies, and reels are challenging in general. 



\begin{table}[h]
    \centering
    \caption{Computational complexities for all models are compared based on number of parameters, memory requirements in Mega-Bytes (MB), and inference speeds in FPS. All experiments are performed on an Nvidia\texttrademark~A$100$ GPU server with $16$\,GB RAM.}
    \vspace{-1mm}
    \begin{tabular}{l|r|r|r}
    \Xhline{2\arrayrulewidth}
        \textbf{Method} & \textbf{\# Params} $\downarrow$ & \textbf{Memory} $\downarrow$ & \textbf{Speed} $\uparrow$ \\
         \Xhline{2\arrayrulewidth}
         FastFCN & $66$\,M & $548.32$\,MB & $16.89$\,FPS   \\
          DeepLabV3+  & $42$\,M & $489.38$\,MB & $15.04$\,FPS  \\
         Segmenter  & $98$\,M & $798.04$\,MB & $13.73$\,FPS \\
         Segformer & $82$\,M & $956.62$\,MB & $10.92$\,FPS  \\
         Swin Tx  & $120$\,M & $1380.16$\,MB & $12.39$\,FPS  \\
         \textbf{CaveSeg} & $\textbf{35}$\,M & $\textbf{406.40}$\,MB & $\textbf{19.78}$\,FPS \\
         \Xhline{2\arrayrulewidth}
    \end{tabular}
    \label{tab:efficiency}
    \vspace{-8mm}
\end{table}

\vspace{2mm}
\subsection{Qualitative Evaluation}
Figure~\ref{fig:comparison} compares the qualitative performance of CaveSeg with other SOTA models. 
A unique feature of CaveSeg and other transformer-based models is that they perform better in finding large and connected areas, \eg, categories such as first/second layer obstacles and open areas. The continuous regions segmented by the CaveSeg model facilitate a better understanding of the surroundings compared to DeepLabV3+ and FastFCN. This validates our intuition that window-shifting technique can indeed extract global features more accurately. While the shifted window method focuses on global feature extraction, the deeper layers and multi-scale pooling of PPM help to preserve the details of local features. The precise detection of caveline, which is only a few pixels wide, further supports this design choice.

 \begin{figure}[b]
     \centering
     {\includegraphics[width=\columnwidth]{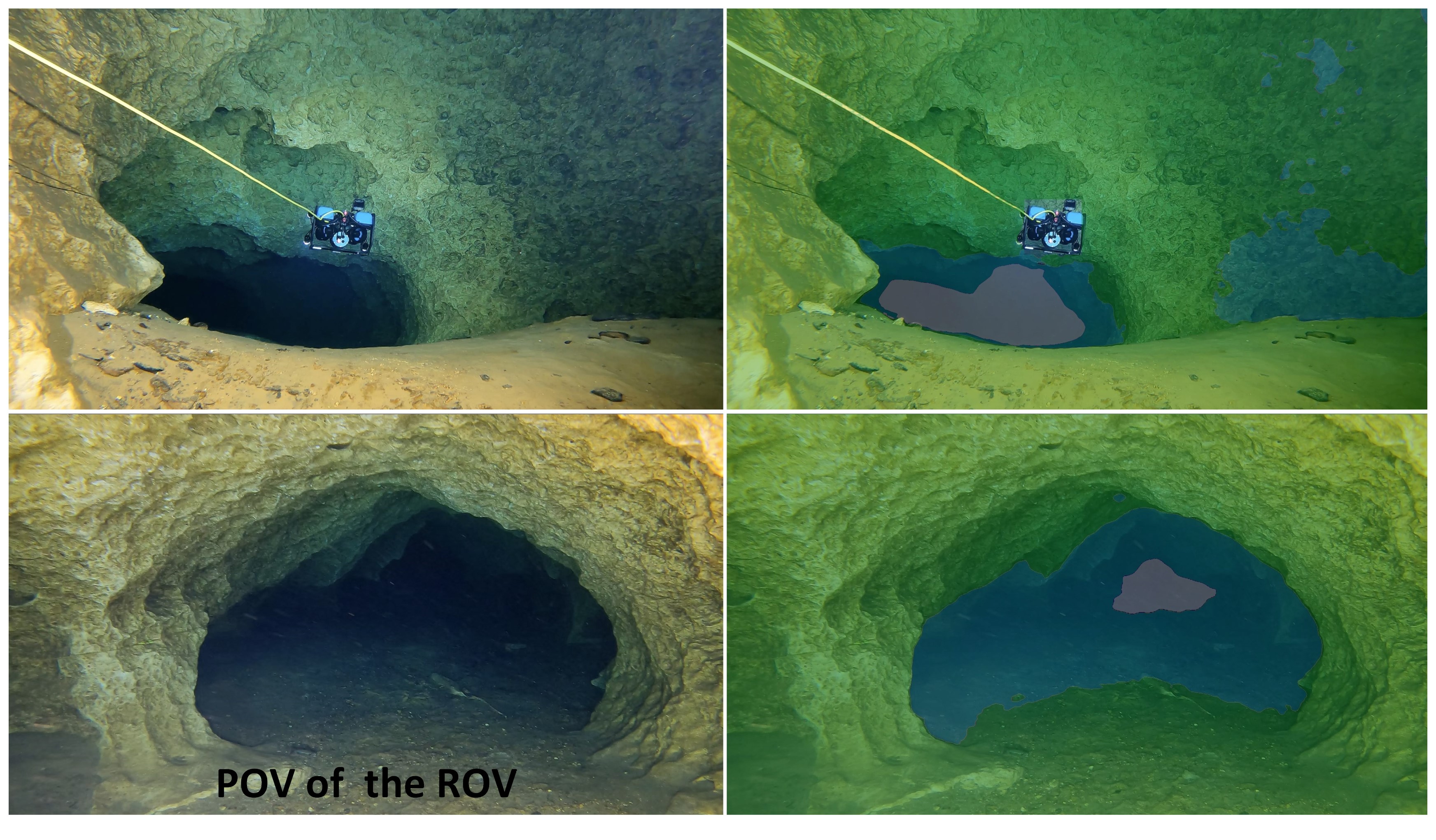}}%
     \vspace{-2mm}
     \caption{A BlueROV2 is being teleoperated through a cave passage in Orange Grove, FL where the caveline is almost invisible; note that the yellow wire on the top image is the ROV's tether, not a caveline. In such scenarios, an AUV can leverage the segmented obstacles and open areas in the scene (overlayed on the right column) toward planning a trajectory to rediscover and follow the caveline.}
     \label{fig:rov_nav}
     \vspace{-2mm}
 \end{figure}

\section{Use Cases: Vision-based Cave Exploration and Semantic Mapping by AUVs}\label{use_cases}
 
\subsection{Safe AUV Navigation Inside Underwater Caves}
\begin{figure}[t]
     \centering
     {\includegraphics[width=\columnwidth]{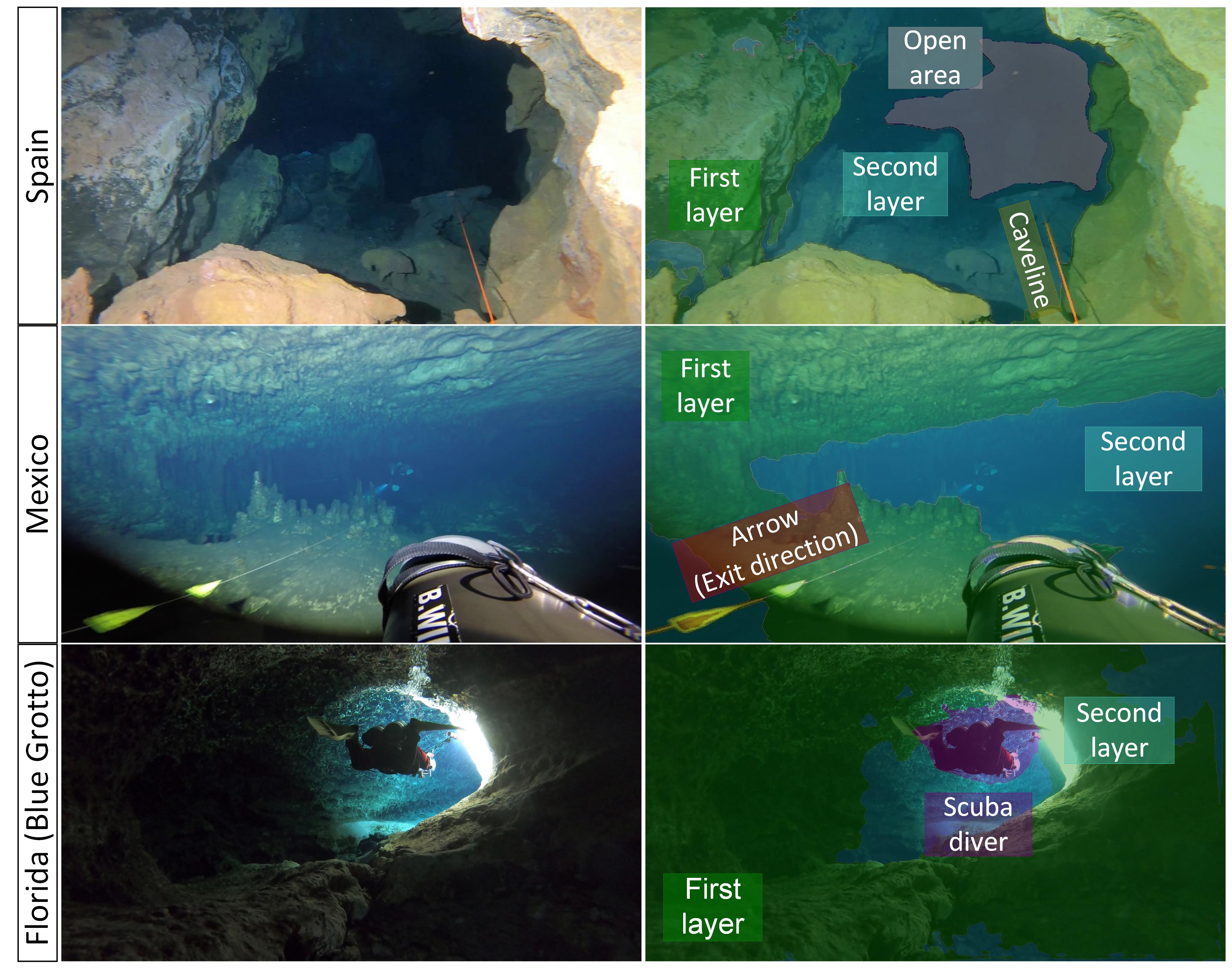}}%
     \vspace{-1mm}
     \caption{A few examples of semantic scene parsing by CaveSeg for important navigation information are shown. The first scene shows a traceable caveline towards the open area and surrounding obstacles. When no open areas are visible (second image), arrows on the caveline are helpful to know the exit direction of the cave. Lastly, the accurate detection of human divers is a crucial step 
     for giving the right-of-way as well as diver-robot cooperation.}
     \label{fig:prac}
     \vspace{-2mm}
 \end{figure}

The confined nature of the underwater cave environment makes obstacle avoidance a particularly important task. Safe navigation approaches, such as AquaNav proposed by Xanthidis \etal~\cite{XanthidisICRA2020} and especially AquaVis~\cite{XanthidisIROS2021} can generate smooth paths by avoiding obstacles. To this end, having a holistic semantic map of the scene can ensure that the AUV can safely follow the caveline and keep it in the FOV during cave exploration. Our previous work~\cite{yu2023weakly} demonstrates the utility of caveline detection and following for autonomous underwater cave exploration. While caveline detection is paramount, having semantic knowledge about the surrounding objects in the scene is essential to ensure safe AUV navigation. 
As shown in Fig.~\ref{fig:rov_nav}, cavelines can be obscure in particular areas due to occlusions and blends inside the cave passages. Hence, dense semantic information provided by CaveSeg is important to ensure continuous tracking and re-discovery of the caveline and other navigation markers. A few more challenging scenarios are shown  Fig.~\ref{fig:prac}, where simply following the caveline is not sufficient due to the cluttered scene geometry. With the semantic knowledge of first layer and second layer obstacles, caveline, and open areas -- an AUV can plan its trajectory safely and efficiently.


\subsection{Working With and/or Alongside Human Scuba Divers}
\begin{figure}[h]
\vspace{-1mm}
     \centering
     {\includegraphics[width=0.49\columnwidth]{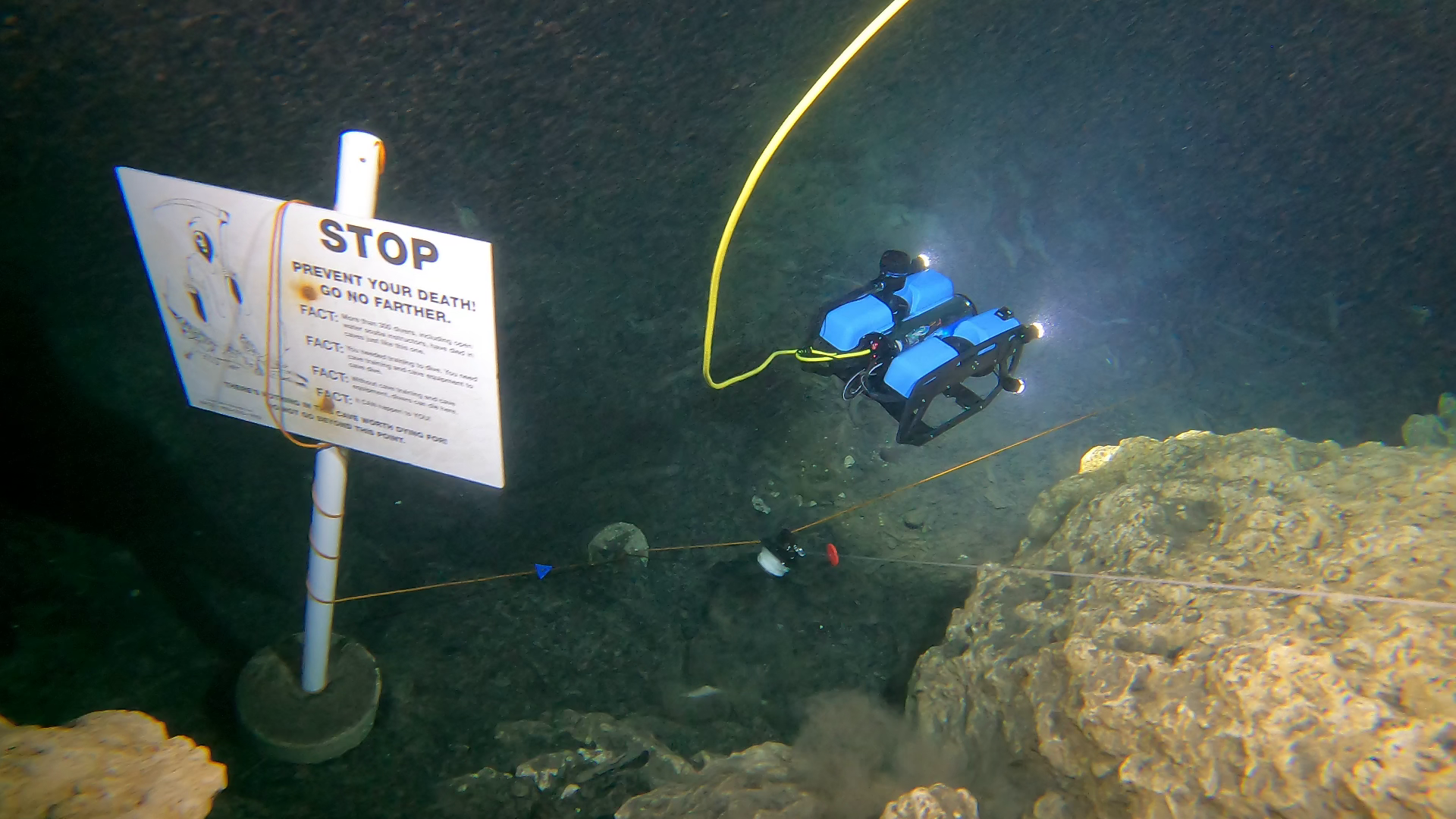}} \label{a}
     {\includegraphics[width=0.49\columnwidth]{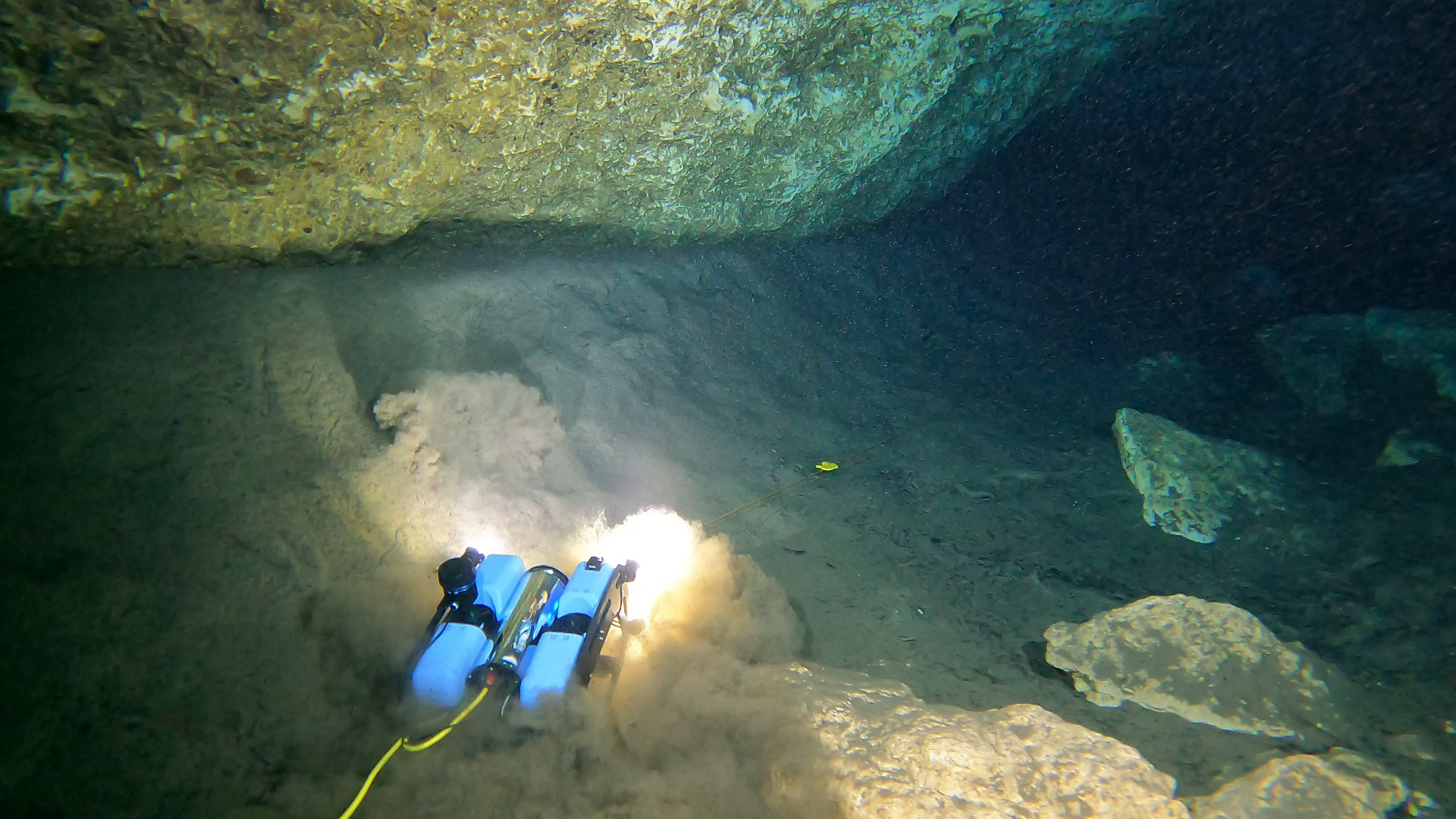}}%
     \label{b}
     \vspace{-1mm}
     \caption{A BluROV2 is being teleoperated inside the Blue Grotto cave, FL: (left) entering the cave; (right) stirring the sediments.}
     \label{fig:Silt}
     \vspace{-5mm}
 \end{figure}

The underwater cave environment is extremely hostile due to the lack of direct access to the surface. Cave divers enforce strict protocols on light configurations (always a primary and two backup lights), use of breathing gas (use only $30\%$ on the way in, leaving $30\%$ for the return and $30\%$ for emergencies), and always keep near the caveline, ensuring there is an uninterrupted line to open water~\cite{exley1986basic}. Of particular importance is the right of way for an exiting team. As such, CaveSeg maintains a label for divers (see Fig.~\ref{fig:prac}) to ensure appropriate actions by the AUV. Specifically, in the presence of a diver, the lights of the vehicle will be dimmed so that the approaching diver can see the caveline. The AUV will reduce its speed, and refrain from using the downward\hyp facing propellers in order to avoid stirring up the sediment; see Fig.~\ref{fig:Silt} where the teleoperated ROV disturbed the sediment on the floor. Finally, the vehicle will move away from the caveline towards the walls of the cave.

\subsection{3D Semantic Mapping and State Estimation}
\begin{figure}[h]
\vspace{-2mm}
     \centering
{\includegraphics[trim={0 1cm 0 2.5cm},clip,width=\linewidth]{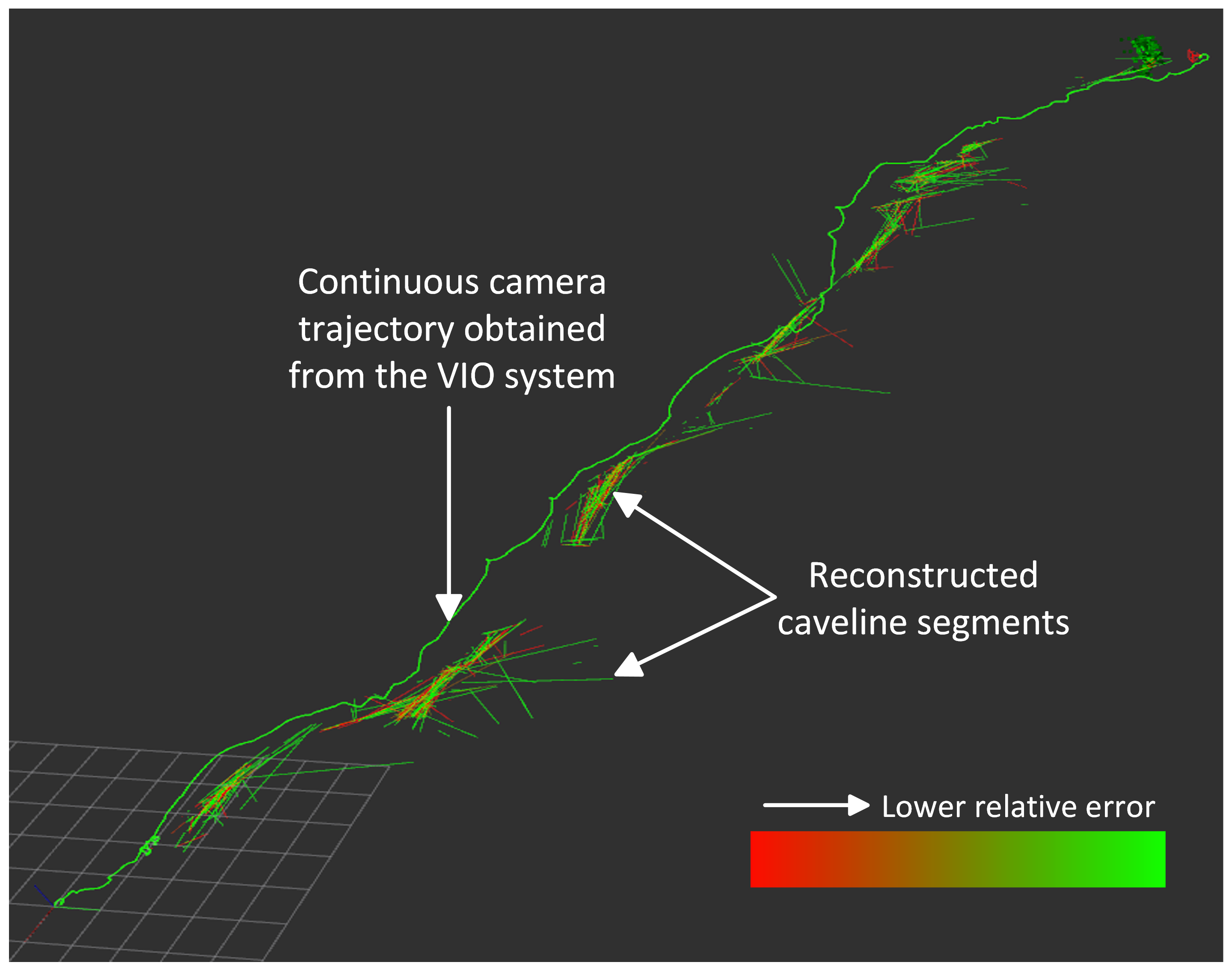}}%
     \vspace{-1mm}
     \caption{Ray-plane triangulation of cavelines using VIO pose estimation and 2D line segments from deep segmentation pipeline are shown. The reconstructed 3D Lines are colored based on average re-projection error within its neighborhood in a local connectivity graph. 
     }
     \label{fig:caveline-reconstruction}
     \vspace{-2mm}
 \end{figure}

Another prominent use case of CaveSeg is the 3D estimation and reconstruction of caveline for AUV navigation. Specifically, 2D caveline estimation can be combined with camera pose estimation from Visual-Inertial-Odometry (VIO) to generate 3D estimates. This could potentially be used to compare and improve manual surveys of existing caves. Moreover, 3D caveline estimations can also be utilized to reduce uncertainty in VIO and SLAM systems since the line directions in 3D point clouds can provide additional spatial constraints~\cite{mo2021fast}. We demonstrate a sample result in Fig.~\ref{fig:caveline-reconstruction}, where \emph{ray-plane triangulation} of cavelines is achieved using Visual-Inertial pose estimation for enabling 3D perception capabilities for underwater cave exploration. This experiment is performed on field data collected by ROVs in the Orange Grove cave system, FL, USA; we are currently exploring these capabilities for more comprehensive 3D semantic mapping of underwater caves.

\section{Conclusion and Future Work}
This paper presents CaveSeg -- the first comprehensive dataset and deep visual learning pipeline for underwater cave scene segmentation. With the primary focus on vision-guided cave exploration and mapping by AUVs, we include semantic labels for object categories such as caveline, layered obstacles, arrows, cookies, attachment points, reels, as well as human scuba divers. We perform extensive benchmark evaluations of several SOTA models that demonstrate the utility of such dataset for robust semantic segmentation of underwater cave scenes. We further propose a computationally light model that offers up to $1.8\times$ faster inference in addition to providing SOTA performance. More importantly, we demonstrate that the predicted scene parsing labels can be utilized for safe AUV navigation inside underwater caves. We are currently investigating the scope of augmenting semantic maps with geometric information from a Visual-Inertial SLAM system such as SVIn2~\cite{RahmanIJRR2022}. This will potentially reveal the 3D position of the caveline as well as the relative location of obstacles and helpful markers. The synergy between geometry and semantics will provide additional visual features to further disambiguate weakly labeled areas during cave exploration. For future CaveSeg releases, we will augment semantic labels for objects such as vertical columns, stalagmite, and stalactite that are found in decorated caves.

\section{Acknowledgements}

This research has been supported in part by the NSF grants $1943205$ and $2024741$. The authors would also like to acknowledge the help of the Woodville Karst Plain Project (WKPP), El Centro Investigador del Sistema Acuífero de Quintana Roo A.C. (CINDAQ),  Global Underwater Explorers (GUE), and Ricardo Constantino, Project Baseline in collecting data, providing access to challenging underwater caves, and mentoring us in underwater cave exploration. We also appreciate the help from Evan Kornacki for coordinating our field experimental setups. We thank the RoboPI lab members (Ruo Chen; Richard Feng) and former students (Ailani Morales; Boxiao Yu) who helped us with image labeling and annotation tasks of CL-ViT and in the early stages of this project.


\printbibliography


\end{document}